\documentclass{easychair}

\usepackage{doc}
\usepackage{graphicx}

\newcommand{\easychair}{\textsf{easychair}}

\title{Engagement Detection in Meetings
\thanks{Empirical data collected in Machine Learning devision at Konica Minolta Laboratory U.S.A.,Inc. The algorithms and methods provided in this research are submitted to United States Patent and Trademark Office (USPTO).}}

\author{
	Maria Frank\inst{1}
\and
	Ghassem Tofighi\inst{2}
\and
	Haisong Gu\inst{3}
\and
	Renate Fruchter\inst{4}	
}

\institute{
  Civil and Environmental Eng., Stanford University,\\ Stanford, California, U.S.A.\\
  \email{mfrank0@stanford.edu}
\and
   Electrical and Computer Eng., Ryerson University,\\
   Toronto, Ontario, Canada\\
   \email{gtofighi@ryerson.ca}\\
\and
   Machine Learning Devision, Konica Minolta Laboratory U.S.A.,Inc\\
   San Mateo, California, U.S.A. \\
   \email{haisong.gu@hl.konicaminolta.us}\\
\and
   Project Based Learning Laboratory (PBL Lab),\\
   Civil and Environmental Eng., Stanford University,\\ Stanford, California, U.S.A.\\
   \email{fruchter@stanford.edu}
 }

\authorrunning{Mokhov, Sutcliffe and Voronkov}

\titlerunning{The {\easychair} Class File}

\begin{document}

\maketitle

\begin{abstract}
 Group meetings are frequent business events aimed to develop and conduct project work, such as Big Room design and construction project meetings. To be effective in these meetings, participants need to have an engaged mental state. The mental state of participants however, is hidden from other participants, and thereby difficult to evaluate. Mental state is understood as an inner process of thinking and feeling, that is formed of a conglomerate of mental representations and propositional attitudes. There is a need to create transparency of these hidden states to understand, evaluate and influence them. Facilitators need to evaluate the meeting situation and adjust for higher engagement and productivity. This paper presents a framework that defines a spectrum of engagement states and an array of classifiers aimed to detect the engagement state of participants in real time. The Engagement Framework integrates multi-modal information from 2D and 3D imaging and sound. Engagement is detected and evaluated at participants and aggregated at group level. We use empirical data collected at the lab of Konica Minolta, Inc. to test initial applications of this framework. The paper presents examples of the tested engagement classifiers, which are based on research in psychology, communication, and human computer interaction. Their accuracy is illustrated in dyadic interaction for engagement detection. In closing we discuss the potential extension to complex group collaboration settings and future feedback implementations.
\end{abstract}


%
%

\pagestyle{empty}

\section{Introduction}
\label{sect:introduction}

Project meetings, such as Big Room design and construction project sessions (Khanzode \& Lamb, 2012 \cite{khanzode2012transcending}), are communicative events where participants discuss, negotiate, present, and create material jointly. Participants express varying mental states in these group meetings due to individual factors and predispositions, the specific meeting agenda, and general team dynamics. To improve meeting quality and productivity it is important to understand the individual user states and the aggregated group state and through this the group dynamics. One characteristic associated with meeting quality, productivity, and effectiveness is engagement.

This paper focuses on a multi-modal approach to detect engagement. Engagement is defined as: “the value that a participant in an interaction attributes to the goal of being together with the other participant(s) and continuing interaction” (Salam, Chetouani 2015, p. 3\cite{salam2015multi}) “the process by which two (or more) participants establish, maintain, and end their perceived connection. It is directly related to attention” (Salam, Chetouani 2015, p. 3\cite{salam2015multi}). It can furthermore be defined as a meaningful involvement (Fletcher 2005\cite{fletcher2012meaningful}), that is enabled through vigor, dedication, and absorption (Schaufeli 2006\cite{schaufeli2006measurement}). For the present study we define engagement as an attentive state of listening, observing, and giving feedback, leading into protagonistic action in group interaction. Individual and group engagement level influences productivity of group interaction. Hence, it is of interest to create transparency about engagement states of participants and group.

An important aspect is the intention to become protagonist and the identification of the threshold between active and passive participation. This moment is crucial because it indicates a significant change in mental state of an individual. Intention is understood as instructions that people give to themselves to behave in certain ways (Triandis 1980\cite{triandis1979values}), and one of the most important predictor of a person’s behavior (Sheeran 2002\cite{sheeran2002intention}). It is an internal commitment to perform an action while in a certain mental state (Levesque et al. 1990\cite{cohen1995intention}) and it predicts actual
behavior. This paper defines intention as a mental state characterized by an internal commitment to perform a specific action in the immediate future. It is a specific form of engagement that is indicating future active meeting participation.

We build on an early study and preliminary prototype, called eRing (engagement Ring), at the PBL Lab at Stanford that focused on detecting and building awareness of learners’ degrees of engagement during globally distributed project team meetings (Ma \& Fruchter, 2015\cite{ma2015ering}). Collocation offers a wealth of tacit cues about learners and team members. Body position and gestures, as well as physical movement provide indicators of participant’s intentions (Pease \& Pease, 2004\cite{pease2008definitive}). These are cues that enable us to intuitively interpret and evaluate the state of learners and team members.

Integrated Delivery Process (IPD) is one of the disruptive forces that the design and construction industry is experiencing. This process leverages Big Room weekly or bi-weekly collocated project stakeholder meetings. During Big Room sessions a lot of time is spent talking in large groups about issues, displaying side by side on SMARTBoards 3D BIM models of the future facility and the Navisworks clashes that need to be resolved (Khanzode \& Lamb, 2012\cite{khanzode2012transcending}). Big Room IPD increases the number of iterations and time-to-market (Fruchter \& Ivanov, 2011\cite{fruchter2011agile}). In order to achieve this increased productivity participant’s need to engage and get feedback on their engagement state.

This paper presents a more comprehensive Engagement Framework that builds on and expands the eRing engagement concept and provides guidelines and instructions for the implementation of a multi-modal sensor-based work environment to detect engagement of participants in a collocated setting. In the preliminary eRing prototype the initial focus was on body motion and posture. The proposed Engagement Framework presented in this paper expands this approach to include other information streams such as facial expression, voice, and other biometric information. We demonstrate the accuracy of the Engagement Framework with a simple interaction scenario, and provide a direction for future implementation.

\section{Theoretical Points of Departure}
\label{sect:theoretical}
The analysis of engagement has a long history in education and work. Work engagement (Schaufeli 2006\cite{schaufeli2006measurement}) influences project outcome. Qualitative observation-based evaluation of an individual’s state has shed light on the importance of engagement for productivity. The Human-Computer-Interaction community started focusing on the automation of engagement detection for learning and games. To combine both methods of engagement detection, qualitative and automated, enables us to provide new insights into automated work engagement detection in the work environment.

A study by Frank and Fruchter (2015\cite{frank2015emotional}) uses cognitive flexibility as an internal evaluation of engagement during meetings. However, this approach as well as self-reports require active involvement of meeting participants. This paper focuses on external expressions of engagement that allow a non-intrusive evaluation of the participant without disrupting the activity. Research in this field focuses on the human body and its features to generate indicators of mental states. Non-verbal behavior and body language expresses emotions (Schindler et al. 2008, Mead 2011\cite{mead2011proxemic}); it conveys information about the speaker's internal state, and attitude toward the addressees (Gutwin \& Penner 2002\cite{gutwin2002improving}). Both body posture and gesture (Ekman 1999\cite{ekman2004emotional}, Ekman and Friesen 1972\cite{ekman1972hand}) have been identified as important social signals that indicate mental states and thereby also engagement. We use the preexisting qualitative research to develop an Engagement Framework offering quantitative measureable indicators.

Various approaches have been investigated to use body language to improve interactions with digital agents such as assistant robots and displays (Vaufrydaz 2015\cite{vaufreydaz2016starting}, Schwarz 2014\cite{schwarz2014combining}). These studies use varying aspects of non-verbal language, such as upper body pose (Mead 2011\cite{mead2011proxemic}, Schwarz 2014\cite{schwarz2014combining}), proxemics (Vaufreydaz 2015\cite{vaufreydaz2016starting}), facial expression and head posture (Vaufreydaz 2015\cite{vaufreydaz2016starting}), or the absence of movement (Witchel 2014 \cite{witchel2014does}) and sometimes the combination of a small set of features from facial expression and posture in multi-modal approaches.

Weight and weight distribution on a seat has been used to measure engagement (Mota \& Picard 2003\cite{mota2003automated}, DeMello et al. 2007\cite{d2007posture}, Balaban 2004 \cite{balaban2004postural}). Weight is an indicator of body posture that can be measured without affecting the participant.

The aforementioned eRing prototype (Ma \& Fruchter 2015 \cite{ma2015ering} focuses on globally distributed learning settings which strip away the physical cues. The hypothesis of the eRing study was that these cues can be reintroduced in virtual interactions, allowing learners to self-regulate and build an awareness of the overall state of the team. The PBL Lab researchers developed a cloud service and application called eRing (engagement Ring) to detect and provide real-time feedback of learner’s degree of engagement during a virtual interaction. eRing collects body motion data using Microsoft Kinect sensor, analyzes and interprets a small set of the body motions and body positions including head, shoulder, and hand joints. eRing provides real-time feedback of the degree of engagement based on three
states – disengaged, neutral listen, and engaged. The body motion analysis is performed for two units of analysis – individual learner and team. eRing runs on a Microsoft Azure cloud platform. eRing was used in the architecture, engineering, construction (AEC) Global Teamwork course testbed in 2014 and 2015.

However, the afore mentioned studies discuss limited sets of potential engagement indicators and do not make full usage of the amount of qualitative research conducted on non-verbal body language and its indication of mental states. Hence, they all provide limited insight and accuracy in evaluating the intention to engage with an agent. They do not discuss the evaluation of engagement between humans in normal interaction or insights into potentially different levels of engagement and thereby provide only a binary classification. This paper discusses a more comprehensive analysis of engagement.

\section{Proposed Engagement Classification Framework}
\label{sect:proposed-engagement}
\begin{figure}[h!]
	\centering
	\includegraphics[width=1.0\textwidth]{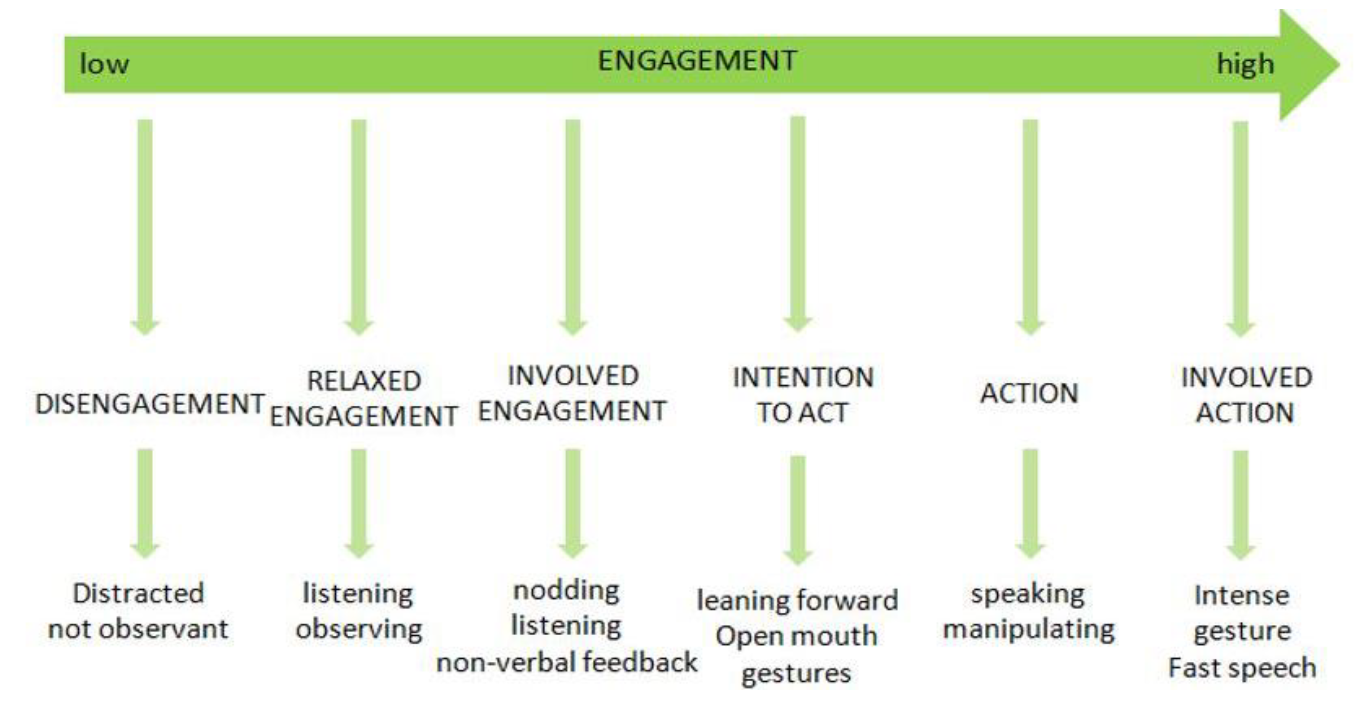}
	\caption{Engagement Framework and sequence of states}
	\label{fig:engagement-framework}
\end{figure}

Building on previous studies, we developed an Engagement Framework in the form of an engagement scale for interaction. Engagement is the level of participation in a meeting interaction. Schwarz et al. (2014)\cite{schwarz2014combining} define two states of disengaged and intention to interact. The eRing prototype defines three states, disengaged, neutral listen, and engaged. The proposed Engagement Framework consists of six states to understand engagement in more detail. Figure \ref{fig:engagement-framework} presents the Engagement Framework consisting of six states. The first state is disengagement; similar to Schwarz et al. (2014)\cite{schwarz2014combining} and Ma and Fruchter (2015)\cite{ma2015ering} disengagement is understood as a state of no participation, distraction and no attention to the meeting. The second state is relaxed engagement, understood as attention to the meeting, listening, observing, but no participation. This is most similar to the neutral listen state of eRing. The third state is involved engagement it signifies attention and non-verbal feedback like nodding, small verbal confirmations, or specific facial expressions such as an open mouth. This state is the first that can be classified as engaged in the eRing prototype, but counts as disengaged in the Schwarz et al. discussion. The present framework differentiates between relaxed and involved engagement because they signify differing mental states and different focus and involvement to the meeting activity and thereby potential for productivity. The fourth state is the intention to act which is similar to the Schwarz model. It is the preparation for active participation in protagonist role indicated through and increase in activity. It shows an involved participation but not yet a contribution. The fifth state is action and the process of speaking and/ or interacting with participants or content on displays. It is a calm form of protagonistic activity. The sixth and final state is involved action and a highly engaged and involved interaction with intense gesture and voice. This state is an active form of excitement and arousal in a meeting that fosters productivity. The six states help understand different team dynamics and user states that affect the meeting interaction, productivity and effectiveness. Using multi-modal information our framework identifies specific combination of engagement classifiers associated with the six engagement states.

\section{Engagement Classifiers}
Literature identified specific observable patterns that indicate mental states such as engagement. In order to categorize the engagement of an individual into one of the six engagement states of the Engagement Framework we need these specific observable patterns. They elucidate which state a participant is in at any given type.

The Engagement Framework draws on a variety of these identified observable patterns understood as indicators for engagement (Sanghvi et al. 2011\cite{sanghvi2011automatic}, Vaufrydaz et al. 2015\cite{vaufreydaz2016starting}, Mead et al. 2011\cite{mead2011proxemic}, Bassetti 2015\cite{bassetti2016social}, Changing minds 2015, Schwarz 2014\cite{schwarz2014combining}, Scherer 2011\cite{scherer2012generic}). As humans we draw on a multitude of indicates to evaluate conversation partner’s engagement. For this reason, the Engagement Framework seeks to implement many identified indicators of engagement to make a more accurate prediction of the engagement state. The indicators are multi-modal, based on body posture and motion, voice, and facial expression to create a more comprehensive perspective. The multi-modal indicators can be captured through ubiquitous sensors such as 3D cameras like Kinect, 2D imaging through RGB cameras, and microphones. The input of each sensor and each mode (3D, 2D, sound) is treated as a module. The module conceptualization allows for future expansion of sensors and indicators to be integrated into the classification of the Engagement Framework.

The indicators are abstracted into classifiers. Classifiers are specific components of body posture or motion, facial expression, voice, and other multi-modal information that can characterize engagement. The classifiers are based on heuristics. To date, we identified 25 indicators, such as volume of speech, smiling, and shaking the head, producing a range of classifiers. Participants in meetings are normally seated behind a table. Hence, the indicators are focused on the upper body with a differentiation between head, face, arms, body, and voice. Table \ref{tab:params} shows an example of indicators and their classifiers for the Body. It names the indicator, its definition, the module, and the classifier in each of the six engagement states. Classifiers are understood as being exhibited or not exhibited during an observation period and are expressed as a binary value. For example, the body lean angle is an indicator of engagement that contains six classifiers. The first classifier is a backwards lean angle that can indicate disengagement. The second classifier is a sideways lean angle that indicates relaxed engagement. The third classifier is no lean angle or sitting straight which shows an involved engagement. The fourth classifier is the process of leaning forward indicating the intention to act. The fifth classifier is a slight forward lean indication action. And the sixth classifier is a rapid change in lean angles over the observation time indicating involved action.

\begin{table}[h]
\centering 	
\caption{Example of classifier selection and definition for the Body}
\begin{tabular}{l c r}
	 	 \includegraphics[width=1.0\textwidth]{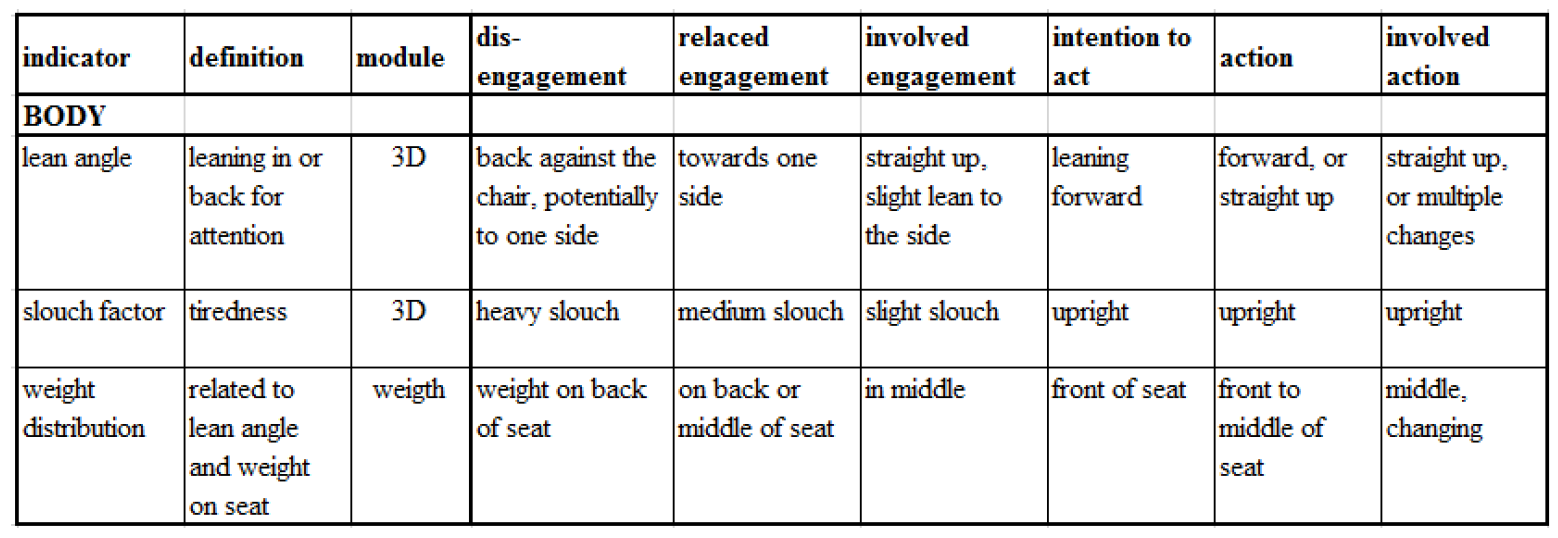}
	 	 &
	 	 &
\end{tabular}
\label{tab:params}
\end{table}

\section{Implementation}
Schwarz et al. (2014\cite{schwarz2014combining}) used a linear regression to calculate weight factors to evaluate the relative importance of five binary classifiers and defined a threshold for the sum of the classifiers to evaluate intention to act. This method is not a valid statistical approach. We propose the implementation of the classifiers as a feature vector containing the binary values. Depending on the available sensors the feature vector consists of features from different modules. The feature vector is fed to a Support Vector Machine with a linear kernel as a machine learning algorithm to evaluate the important classifiers and cluster the emerging pattern in the Engagement Framework states. Our approach is based on LIBSVM. LIBSVM implements the Sequential minimal optimization (SMO) algorithm for kernelized SVMs, supporting classification and regression.

\begin{figure}[h!]
	\centering
	\includegraphics[width=1.0\textwidth]{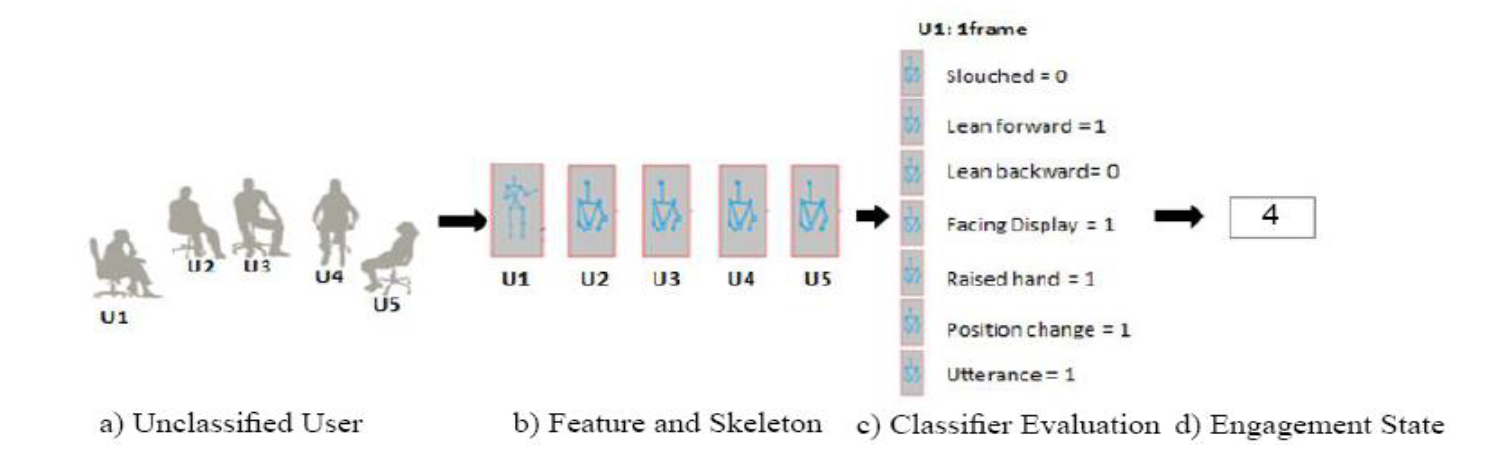}
	\caption{Implementation of Engagement Framework and evaluation of engagement state}
	\label{fig:implementation}
\end{figure}

Figure \ref{fig:implementation}  illustrates the first stage of an example implementation. Each meeting includes various team members. For the implementation system these team members are unclassified (Figure \ref{fig:implementation}a). Each individual meeting participant is represented through one feature abstraction, which is a skeleton 3D sensor abstraction in this example (Figure \ref{fig:implementation}b). Based on the skeleton the related set of binary classifiers is instantiated and creates the feature vector for each team member (Figure \ref{fig:implementation}c). The SVM then classifies the individual engagement state from one to six (disengagement to involved action) accordingly (Figure \ref{fig:implementation}d).

\begin{figure}[h!]
	\centering
	\includegraphics[width=0.4\textwidth]{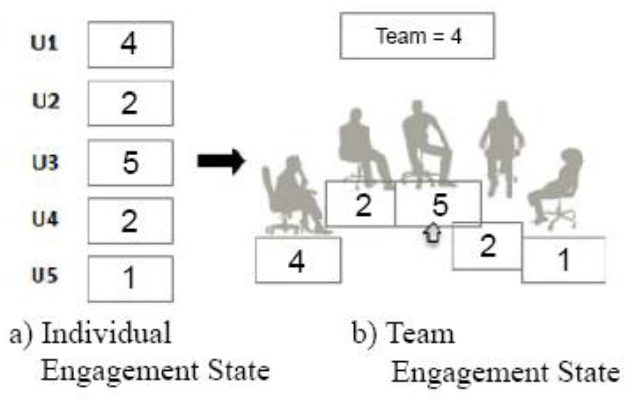}
	\caption{Classification of individual and team engagement state}
	\label{fig:classification-ind}
\end{figure}

Figure \ref{fig:classification-ind}  illustrated the final team engagement classification. The Engagement Framework Implementation creates multiple feature vectors to evaluate every participant in a meeting situation and classify their indicators into one of the six engagement states \ref{fig:classification-ind}a). The classification result for each team member is then aggregated to show the engagement state distribution of the team and calculate the average engagement state as team state (Figure \ref{fig:classification-ind}b).

\section{Testing}
\subsection{Scenario}
\begin{figure}[h!]
	\centering
	\includegraphics[width=1.0\textwidth]{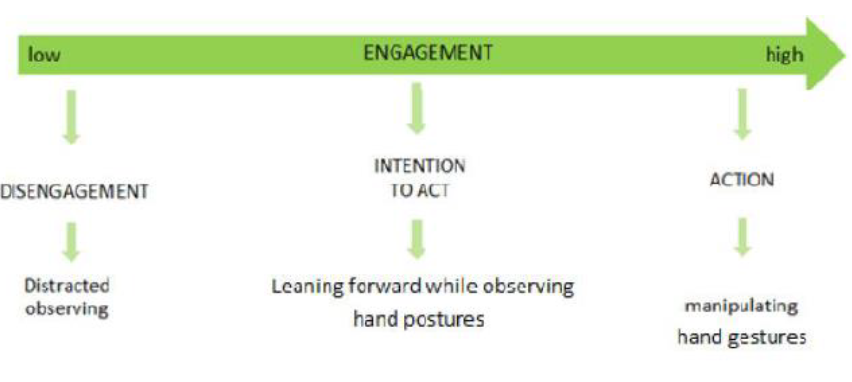}
	\caption{States implemented in initial experiment}
	\label{fig:states-implemented}
\end{figure}

In an initial experiment a dyad is observed to train the algorithm and test prediction accuracy.
The focus of this experiment is, similar to eRing (Ma \& Fruchter 2015\cite{ma2015ering}) on three out of the six states: disengagement, intention to act, action (Figure \ref{fig:states-implemented}). We implemented the body motion and posture module based on 3D joint information for this experiment. Due to the complexity of the Engagement Framework, and the implementation through multi-modal modules and classifiers we had to simplify the first experiment as a starting point. This simplification allows for an initial engagement classification and testing of the implementation of the classifiers.

\subsection{Data Collection}
Schwarz (2014\cite{schwarz2014combining}) introduced a simple task hand-off game to evaluate intention to interact. We implemented a similar task hand-off game to observe and detect the three engagement states of disengagement, intention to act, and action and create a clear ground truth. The dyad’s task is to select colored blocks on a display by using a hand to navigate a cursor. Colors of the blocks change for different participants. When participant 1 is playing the blocks are green. The blocks turn red when participant 2 is playing. The task starts with a 10 second countdown, participant 1 then has 10 seconds toselect as many blocks as possible. There is a 10 second switch period for participant 2 to get ready and have the intention to act. Participants are penalized for not disengaging after their 10 second segment. The game runs for 20 total switches. Each participant group plays the game 5 times. Participants were positioned in about 2.5 m distance from a screen. A depth-sensing camera Kinect (Xbox one) is installed underneath the screen to capture body motion.

Data was collected from six members of the research team at Konica Minolta: 3 female, 3 male, all are professionals in IT and have a moderate to high experience with motion detection. The participants are randomly assembled in pairs to play the game.

\subsection{Implemented Classifiers}
\begin{figure}[h!]
	\centering
	\includegraphics[width=0.4\textwidth]{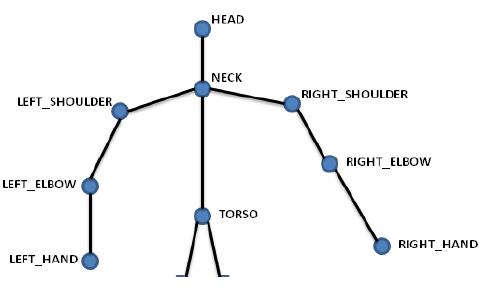}
	\caption{States implemented in initial experiment}
	\label{fig:3d-joints}
\end{figure}

The 3D sensor module using the ASUS Xtion, used the data of the upper body joints (Figure \ref{fig:3d-joints}). Participants are normally seated behind a table in meetings. Therefore, only the body joints for head, left and right shoulders, left and right elbows, left and right hand, torso and left and right hips are relevant.

Table \ref{tab:classifiers} shows the list of binary classifiers implemented in the initial experiment. A total of 5 indicators and their 16 classifiers is used to create the feature vector. The classifiers are evaluated on a frame by frame basis. This means that for each frame of the task recording a feature vector of the binary classifiers is created and the engagement state classified.

\begin{table}[h]
	\centering 	
	\caption{Classifiers of initial experiment}
	\begin{tabular}{l c r}
		\includegraphics[width=1.0\textwidth]{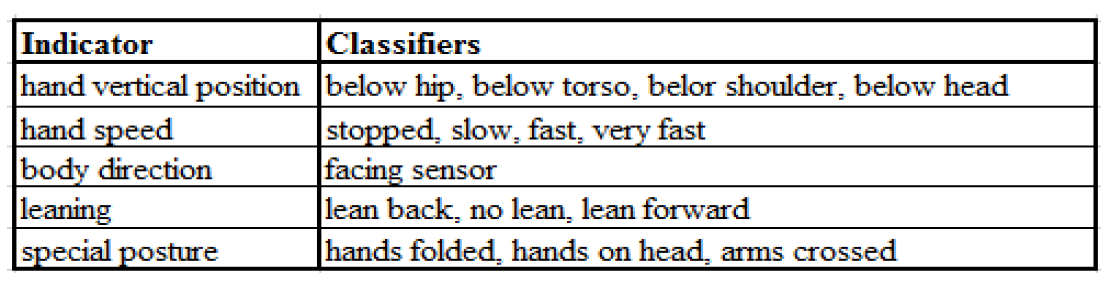}
		&
		&
	\end{tabular}
	\label{tab:classifiers}
\end{table}

We captured and labeled 2,321 frames from 6 different subjects for Intention to Act and Disengagement. Five hundred frames (about 20\% of the frames) are used for training and the remaining frames are used for testing our classifiers. Each frame is classified as Disengagement or Intention to Act. The Action state occurs when a frame label is Intention to Act and the speed of one hand is not zero. The initial implementation reached a classification accuracy of 83.36\%.

The processing time for each frame is less than 10ms which indicates real-time usability of our algorithm since we receive 30 frames per second and should label them to one of the states of engagement.

\section{Discussion}
Automated engagement detection can assist to increase our understanding of group dynamics, and participation habits. The presented Engagement Framework was implemented using research based observable engagement indicators and classifiers show a high accuracy in predicting participation behavior. This provides the basis for future comprehensive team engagement analyses. Our proposed approach to engagement classification promises a more comprehensive observation of individual engagement and team engagement for various applications.

Throughout the meeting the participation level of participant changes. Calculating the engagement state of individual and group in real time allows for feedback of group activity and enables managerial and environmental adjustment to create a more engaging and thereby productive work environment.

\subsection{Engagement Feedback For Meeting Participant and Infrastructure Management}
The individual and team level engagement state can help to inform the meeting in process on various levels, and we elaborate on some of the application opportunities.

A dynamic meeting feedback can trigger a signal at individual or team level; either targeting the disengaged individual to encourage reengagement or the group to change the meeting focus or activity. Groups would receive a joint message such as “The group is disengaged” or a motion signal is triggered like eRing (Ma \& Fruchter 2015\cite{ma2015ering}), which means that the SVM classified more than 40\% of the individuals as disengaged. The public feedback is used as an encouragement to change the meeting activity or take a break to refocus. Detailed insight is helpful to evaluate meeting effectiveness and efficiency.

Records of engagement levels will provide insight to bigger trends in meeting dynamic, such as overall engagement at specific meeting times, participation, and dominance in meetings. These insights allow for strategic adjustments of meeting times, necessary participants, necessary changes in interaction protocols, and others aspects related to meeting productivity.

\subsection{Operation Intent Feedback}
The engagement level can be used to identify the user intent to act with various responsive objects in a room by identifying directed engagement levels. It will give additional information to responsive objects as to which user wants to interact with which object in a room, based on directional body posture, voice, or motion. It will use an engagement table to track each participants’ potential engagement level with each object.

\subsection{Limitations and Future Research}\label{sect:limitations}
The presented paper is an initial step in developing a complex responsive engagement system. The accuracy of the algorithm can be increased. In this initial stage of the project a limited number of classifiers have been implemented. Currently the study focuses on body posture and motion. The classifiers are designed as modules that allow for expansion through different sensors, integrating for example voice data, and facial expression, up to visions of the smart office future including biometric information. The expansion of the classifier list will allow for a more complex understanding of engagement and user state. Furthermore, on-going efforts focus on the implementation of a machine learning algorithm that will allow for a faster and more accurate classification of the six engagement states. The collection of more training data will further assist is making the Framework implementation more robust.

The test game presented in the paper is an abstraction and simplification of real interactions and hence future experimental scenarios will consider more complex future steps. We will present the teams with increasingly complex interaction tasks up to the implementation of the algorithm in realistic meetings.

The final goal for the implementation of the Engagement Framework is to continuously classify the engagement of multiple meeting participants that are within the operating range of the sensors for all six engagement states. Throughout a meeting participants listen, observe, talk, or interact with a display. The behaviors are evaluated through multi-modal modules of engagement classifiers to categorize engagement into six stages. Such a system will assist to detect and evaluate engagement states, and foster engagement in scenarios ranging from small meeting groups to Big Room project stakeholder meetings.

\label{sect:bib}



\end{document}